\documentclass{article}

\usepackage[final,sglblindworkshop]{neurips_2026}
\workshoptitle{Who Verifies the Agents? Toward Reliable Agent Development}
\usepackage[utf8]{inputenc}
\usepackage[T1]{fontenc}
\usepackage{hyperref}
\usepackage{url}
\usepackage{booktabs}
\usepackage{amsfonts}
\usepackage{amsmath}
\usepackage{amssymb}
\usepackage{microtype}
\usepackage{graphicx}
\usepackage{multirow}
\usepackage{tikz}
\usetikzlibrary{positioning,arrows.meta,decorations.pathreplacing,calc}

\title{Keep Your Receipts:\\ Trustworthy Cleanup of Agent-Built Knowledge Graphs}
\author{%
  Vishisht Choudhary\\
  Technical University of Munich\\
}

\begin{document}
\maketitle
\vspace{-4mm}
\begin{center}\small\itshape
Preprint; the camera-ready for the NeurIPS 2026 workshop ``Who Verifies the Agents?'' may differ.
\end{center}
\vspace{-2mm}

\begin{abstract}
Agents build knowledge graphs as memory, and building them has become easy.
Maintaining them has not. When a source is retracted, overruled, or erased,
everything built on top of it, entity merges, entailed edges, community
summaries, has to be re-examined, and the maintainer must choose between
deleting too much and deleting too little. Deployed pipelines delete only the
directly extracted facts. On our corpus that reduces contaminated queries
from 32.2\% to 28.0\% under a real retraction and leaves predicate
revocation unchanged at 36.8\%. Rebuilding is no repair for a constructor
that samples: two builds of the same 300-document corpus differ in 29{,}475
artifacts, five times the 6{,}248 that removing the 61 retracted documents
changes. PolyGraph instead asks the constructor for receipts: for every
artifact, alternative bundles of source evidence, each claimed sufficient on
its own, emitted in the call that creates the artifact. Cleanup is then set
arithmetic over the receipts, milliseconds on a 24{,}381-artifact graph,
with no model calls. Because receipts are the model's own self-report, a
re-derivation audit checks them by deleting the cited evidence and testing
whether the constructor still produces the artifact. The audit finds that a
receipt is trustworthy when it is shaped like the claim it backs. Merge
receipts, each backing one narrow claim, agree with re-derivation 68\% and
85\% of the time. Summary receipts, elicited per sentence but stored as one
flat list for the whole text, are wrong about two thirds of the time under
the two graded deletions and 41\% of the time pooled over ten further random
retractions, almost always by keeping dead content. Eliciting one receipt
per sentence instead gives receipts that are 100\% and 88\% faithful and
lets cleanup keep 33 of 35 touched summaries by removing only their dead
sentences. The layer ordering and the direction of the errors hold under a
second oracle model, a claim-blind prompt, a second domain and a second
constructor. Distrusting the summary layer cuts contaminated queries to
0.4\% and 0.8\% while retaining 99.8 to 100\% of the recall a perfect
maintainer would return.
\end{abstract}

\section{Introduction}
\label{sec:intro}

Knowledge graphs are becoming the ground truth for agentic memory systems,
and constructing them has become easy. Maintaining them has not. This paper
is about one part of maintenance: removing an invalidated source from an
agent-built graph without also removing knowledge that still stands, and
without rebuilding the graph to find out which is which.

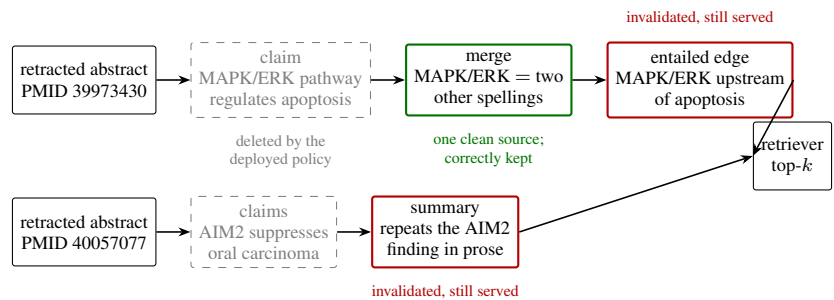
\begin{figure}[t]
\centering
\begin{tikzpicture}[
  node distance=3mm and 4.5mm,
  box/.style={draw, rounded corners=1pt, align=center, font=\scriptsize,
              inner sep=3pt, minimum height=9mm},
  dead/.style={box, draw=gray, text=gray, dashed},
  keep/.style={box, draw=green!45!black, thick},
  live/.style={box, draw=red!70!black, thick},
  lab/.style={font=\tiny, align=center},
  arr/.style={-{Stealth[length=1.6mm]}, semithick}
]
\node[box] (doc1) {retracted abstract\\PMID 39973430};
\node[dead, right=of doc1] (claim1) {claim\\MAPK/ERK pathway\\regulates apoptosis};
\node[keep, right=of claim1] (merge) {merge\\MAPK/ERK $=$ two\\other spellings};
\node[live, right=of merge] (edge) {entailed edge\\MAPK/ERK upstream\\of apoptosis};
\node[box, below=11mm of doc1] (doc2) {retracted abstract\\PMID 40057077};
\node[dead, right=of doc2] (claim2) {claims\\AIM2 suppresses\\oral carcinoma};
\node[live, right=of claim2] (sum) {summary\\repeats the AIM2\\finding in prose};
\node[box] (ret) at ($(edge.east)!0.5!(sum.east) + (18mm,0)$) {retriever\\top-$k$};
\draw[arr] (doc1) -- (claim1);
\draw[arr] (claim1) -- (merge);
\draw[arr] (merge) -- (edge);
\draw[arr] (edge.east) -- (ret.west);
\draw[arr] (doc2) -- (claim2);
\draw[arr] (claim2) -- (sum);
\draw[arr] (sum.east) -- (ret.west);
\node[lab, below=1mm of claim1, text=gray] {deleted by the\\deployed policy};
\node[lab, below=1mm of merge, text=green!45!black] {one clean source;\\correctly kept};
\node[lab, above=1mm of edge, text=red!70!black] {invalidated, still served};
\node[lab, below=1mm of sum, text=red!70!black] {invalidated, still served};
\end{tikzpicture}
\caption{Two real chains from the corpus. Deployed deletion removes only the
extracted claims (dashed). The entity merge keeps one evidence source from a
paper that still stands, so it is correctly retained (green); five of five
re-derivations agree. The entailed edge and the summary rest only on
retracted evidence, mention no retracted paper, and keep reaching the
retriever (red). Support labels delete the edge with one disjointness test.
The summary's label wrongly claims it still stands, the failure
Section~\ref{sec:findings} measures; the layered policy of
Section~\ref{sec:stakes} removes it.}
\label{fig:cascade}
\end{figure}

The difficulty has a simple shape. Two corpora X and Y support an inferred
fact Z. When X is retracted, deleting X's own statements is easy, but Z must
be re-examined: perhaps Y alone justifies it, perhaps not. Once further
deductions are built on Z, each fed by further sources, the cost of deciding
what still stands grows with every layer. Figure~\ref{fig:cascade} shows two
real instances from our corpus. The maintainer has two ways to fail. Deleting everything
downstream of the retraction destroys artifacts that a second, untouched
derivation still supports. Deleting only the directly extracted facts, which
is what deployed pipelines do, leaves most of the contamination in place.

Existing database systems each solve half of the problem. Versioned stores
can rewind to any past state but track no dependencies between items, so a
retraction can be undone but never propagated. The classical propagation
algorithms \citep{gupta1993dred,dekleer1986atms,green2007provenance} would
cascade it exactly, but they consume a justification record that only a
deterministic engine can write; a language model does not produce one, and
re-running the model cannot recover it (Section~\ref{sec:replay}). The
pipelines that actually build agent graphs
\citep{edge2024graphrag,guo2024lightrag} store provenance and use it for
nothing beyond deleting the directly extracted facts. On our corpus that
cleanup reduces contaminated queries from 32.2\% to 28.0\% under a real
retraction, about one eighth of the problem, and under predicate revocation
it changes nothing, 36.8\% before and after, because the damage sits in
derived edges and summaries rather than in the facts.

PolyGraph closes the gap with two steps. First, the constructor must emit a
receipt for every artifact in the same call that creates it. A receipt is a
support label \citep{dekleer1986atms}: a set of alternative evidence
bundles over source assertions, each claimed sufficient on its own. Cleanup
then needs no model calls, because the information that removal requires
was stored before it was needed; processing a withdrawal is set arithmetic
over the receipts, 1 to 21\,ms on our whole 24{,}381-artifact graph.
Second, because the receipts are the model's own self-report, a
re-derivation audit grades them: it deletes the cited evidence and asks
whether the constructor, shown only what survives, still produces the
artifact. It is important to keep your receipts, and just as important to
check who wrote them.

The audit's verdict is that trust depends on the shape of the receipt. At
the merge layer, where each receipt backs one narrow identity claim, the
receipts are trustworthy. At the summary layer, where one flat receipt has
to vouch for several sentences, they are wrong about two thirds of the
time, almost always by keeping content that no longer re-derives. A
controlled re-elicitation traces most of that failure to the shape of the
receipt rather than to the model: asked for one receipt per sentence, the
same constructor writes receipts that are 100\% and 88\% faithful, and
cleanup can keep the sentences that still stand instead of deleting whole
summaries. The general lesson is that a receipt is trustworthy when it is
shaped like the claim it backs, so an agent's outputs should decompose into
narrow, falsifiable sub-claims, each with its own receipt.

The paper contributes:
\begin{itemize}\setlength\itemsep{0pt}
\item the receipt-and-audit protocol, with a distinction between the
  liveness the audit measures and the source-groundedness a compliance
  setting would need (Section~\ref{sec:protocol});
\item the measurement that replay cannot stand in for it on a sampling
  constructor: resampling noise is five times the signal a deletion leaves
  (Section~\ref{sec:replay});
\item the layer-dependent faithfulness result, its receipt-shape
  explanation, and the sentence-level receipts that repair it
  (Sections~\ref{sec:layer} to~\ref{sec:shape});
\item evidence that the result is not an artifact of the sample, the
  oracle, the corpus, or the constructor: ten further pooled deletions, a
  second oracle model and a claim-blind prompt, a second domain (Supreme
  Court opinions) and a second constructor
  (Section~\ref{sec:robust});
\item the retrieval-level consequence: acting on the verdict cuts
  contaminated search results from 32 to 37\% of queries to 0.4\% and
  0.8\% while keeping 99.8 to 100\% of the recall a perfect maintainer
  would return, and what the receipts cost to write and store
  (Section~\ref{sec:stakes}).
\end{itemize}

\section{Related work}
\label{sec:related}

\textbf{Faithfulness of model self-reports.} Chain-of-thought explanations
can misrepresent the causes of a model's answer \citep{turpin2023unfaithful},
and perturbation studies find reasoning traces whose removal leaves the
answer unchanged \citep{lanham2023measuring}. Our self-report differs in one
useful way: it is not free text but a set of support sets with defined
deletion semantics. Each label makes a falsifiable prediction about a
counterfactual corpus, and re-derivation runs that counterfactual.

\textbf{Verification by a second model.} LLM-as-judge protocols
\citep{zheng2023llmjudge} and RAG evaluation suites \citep{es2024ragas} score
outputs with another model call. For our problem this reintroduces the
circularity. A judge asked whether some evidence supports an artifact
answers a different question from the one maintenance needs, namely whether
the deployed constructor would still have produced the artifact from the
surviving evidence. Re-derivation puts that question to the constructor
itself; Section~\ref{sec:robust} measures what a second model and a
judge-style prompt change.

\textbf{Justification-consuming maintenance.} Truth-maintenance systems
\citep{doyle1979tms,dekleer1986atms}, delete-and-rederive
\citep{gupta1993dred}, and provenance semirings \citep{green2007provenance}
settled how to propagate retractions given correct justifications; even
then, minimal deletion is NP-hard \citep{buneman2002deletion}. Versioned
stores such as TerminusDB \citep{vanotterdijk2020succinct} rewind exactly
but track no dependencies between items. All of this machinery assumes that
recorded antecedents are causal and that re-running the producer replays
them. We quantify how badly the replay assumption fails for a language-model
producer, and how unevenly the antecedent assumption holds.

\textbf{Unlearning.} Machine unlearning removes training data's influence
from weights \citep{bourtoule2021unlearning,meng2022rome}, with extensions
to RAG corpora \citep{wang2024ragunlearning}. Our facts live in an
inspectable artifact the model built, not in weights, and the
retraining-style remedy is the replay that Section~\ref{sec:replay} rules
out.

\section{Setting}
\label{sec:setting}

We separate what a document says from what the graph holds. An
\textbf{assertion} $x = \langle d, s, p, o \rangle$ is one source document
$d$ stating one typed triple; assertions are what a withdrawal removes. An
\textbf{artifact} is anything the graph holds and a retriever can return:
an extracted claim, an entity merge, a schema-entailed edge, or a community
summary. Each kind is produced by one pipeline stage built on the one
before it (claims from documents, merges over claims, edges entailed from
claims and merges, summaries over communities of merged entities), and we
call each kind a \emph{layer}. The constructor must emit, in the same call that creates artifact
$a$, a \textbf{support label}
\[
  L(a) \;=\; \{\, S_1, \dots, S_m \,\}, \qquad S_i \subseteq \mathcal{X},
\]
where each support set $S_i$ is a set of assertions claimed to suffice, on
its own, to produce $a$. Under a deletion $D$ the label predicts that $a$
survives exactly when some $S_i$ is disjoint from $D$. Labels are capped at
eight support sets, keeping the smallest. Truncation can only make an
artifact look deader than it is, so it never adds leakage.

\textbf{Deletions.} We grade labels under two operations.
\textsc{retract-source} removes every assertion of a withdrawn document.
\textsc{revoke-relation} removes every assertion that uses a predicate
judged unsound, so it cuts along the schema instead of along documents.

\textbf{Graphs.} The main instance is a pipeline over 300 PubMed oncology
abstracts selected by shared MeSH descriptors \citep{nlm_mesh}, 61 of them
genuinely retracted; its \textsc{retract-source} uses those real
retractions. The constructor is Anthropic's Claude Sonnet, called in August
2026 through the Claude Code CLI's \texttt{sonnet} alias behind a
content-addressed cache keyed by model and prompt hash; the re-derivation
oracle later queries the same backend and alias, so labels are graded
against the producer that wrote them. Every later run in this paper pins
\texttt{claude-sonnet-5} or \texttt{claude-opus-5-5}. The prompt templates
for every model-facing step are released verbatim with the code.\footnote{%
\url{https://github.com/VishiATChoudhary/polygraph}} The
pipeline produced 2{,}785 assertions over 2{,}784 distinct claims, 165
entity merges, and 59 community summaries; closed under the published
axioms of the OBO Relations Ontology \citep{obo_ro}, these yield 21{,}373
entailed edges, for 24{,}381 artifacts in total. Two further instances test
transfer (Section~\ref{sec:robust}). The second domain is 150 U.S.\
Supreme Court opinions from the Caselaw Access Project (volumes 380 to 387
of the U.S.\ Reports), built by the same pipeline and prompts under a legal
schema: 1{,}571 assertions, 136 merges, 60 summaries and 6{,}137 entailed
edges. That corpus has no real retractions, so its \textsc{retract-source}
draws documents at random (seed 0) to the same 10\% size, and its queries
are 300 party and citation strings. The second constructor rebuilds the
biomedical graph from the same abstracts, prompts and schema with
\texttt{claude-opus-5-5}.

\section{PolyGraph: the verification protocol}
\label{sec:protocol}

\begin{figure}[t]
\centering
\resizebox{\textwidth}{!}{%
\begin{tikzpicture}[x=1cm,y=1cm,
  story/.style={font=\scriptsize, align=center},
  proto/.style={font=\tiny\itshape, align=center, text=blue!35!black},
  ptitle/.style={font=\scriptsize\bfseries, align=center},
  panel/.style={draw=black!25, rounded corners=2pt},
  arr/.style={-{Stealth[length=2mm]}, thick, black!55}]

\newcommand{\wolfpic}[3]{
\begin{scope}[shift={#1}, scale=#2]
\fill[#3]
 (0,0.42) -- (0.12,0.50) -- (0.22,0.62) -- (0.26,0.78) -- (0.34,0.62)
 -- (0.40,0.78) -- (0.46,0.58) -- (0.70,0.52) -- (1.05,0.55) -- (1.35,0.50)
 -- (1.50,0.40) -- (1.68,0.24) -- (1.58,0.16) -- (1.40,0.28) -- (1.38,0.05)
 -- (1.30,0.05) -- (1.27,0.30) -- (1.12,0.28) -- (0.78,0.30) -- (0.74,0.05)
 -- (0.66,0.05) -- (0.60,0.32) -- (0.30,0.40) -- (0.10,0.36) -- cycle;
\end{scope}}
\newcommand{\bookpic}[4]{
\begin{scope}[shift={#1}, scale=#2]
\draw[fill=#3, draw=black!60] (0,0) rectangle (0.62,0.82);
\draw[black!60] (0.09,0) -- (0.09,0.82);
\node[font=\footnotesize\bfseries] at (0.36,0.41) {#4};
\end{scope}}

\def\pw{3.35}   
\def\ph{3.55}   

\begin{scope}[shift={(0,0)}]
\draw[panel] (0,0) rectangle (\pw,\ph);
\node[ptitle] at (0.5*\pw,3.25) {1\; the claim};
\fill[black!14] (0.30,1.55) -- (0.30,2.45) -- (0.44,2.75) -- (0.60,2.75)
  -- (0.74,2.45) -- (0.74,1.55) -- cycle;
\draw[black!28, line width=0.4pt] (0.34,1.70) -- (0.70,2.40) (0.70,1.70) -- (0.34,2.40);
\fill[black!14] (0.84,1.55) rectangle (0.96,2.60);
\fill[black!14] (2.00,1.55) -- (2.22,1.90) -- (2.62,1.90) -- (2.84,1.55) -- cycle;
\draw[black!32, line width=0.55pt]
 (2.10,1.92) .. controls (2.24,2.38) and (2.40,2.38) .. (2.48,2.14)
 .. controls (2.54,1.96) and (2.40,1.94) .. (2.40,2.10)
 .. controls (2.40,2.30) and (2.62,2.28) .. (2.74,1.96);
\draw[black!30] (0.15,1.55) -- (3.2,1.55);
\wolfpic{(1.05,1.58)}{0.62}{black!70}
\node[story] at (0.5*\pw,1.13) {``Wolves live in Duisburg,\\ by book A or book B.''};
\node[proto] at (0.5*\pw,0.38) {artifact $a$ with support label\\ $L(a)=\{\{A\},\{B\}\}$};
\end{scope}

\begin{scope}[shift={(3.95,0)}]
\draw[panel] (0,0) rectangle (\pw,\ph);
\node[ptitle] at (0.5*\pw,3.25) {2\; a book is pulled};
\bookpic{(0.60,1.70)}{1.15}{red!8}{A}
\bookpic{(1.85,1.70)}{1.15}{green!10}{B}
\draw[red!70!black, line width=1.2pt] (0.52,1.62) -- (1.40,2.72);
\draw[red!70!black, line width=1.2pt] (0.52,2.72) -- (1.40,1.62);
\node[story] at (0.5*\pw,1.13) {book A is retracted;\\ the receipt says: ``B remains,\\ the sentence may stay''};
\node[proto] at (0.5*\pw,0.38) {deletion $D$ hits $\{A\}$;\\ label predicts alive via $\{B\}$};
\end{scope}

\begin{scope}[shift={(7.90,0)}]
\draw[panel] (0,0) rectangle (\pw,\ph);
\node[ptitle] at (0.5*\pw,3.25) {3\; rewrite, five times};
\bookpic{(0.35,1.95)}{0.95}{green!10}{B}
\draw[arr] (1.15,2.30) -- (1.60,2.30);
\foreach \i in {0,...,4}{
  \draw[fill=white, draw=black!55]
    ({1.72+\i*0.24},{1.78-\i*0.05}) rectangle ({2.32+\i*0.24},{2.62-\i*0.05});
}
\node[font=\tiny, text=black!60] at (2.55,2.20) {$\times 5$};
\node[story] at (0.5*\pw,1.13) {take A away, hand over\\ only B: ``write that part\\ of the report again''};
\node[proto] at (0.5*\pw,0.38) {$R=5$ re-derivation draws\\ on $\mathcal{X}\setminus D$};
\end{scope}

\begin{scope}[shift={(11.85,0)}]
\draw[panel] (0,0) rectangle (\pw,\ph);
\node[ptitle] at (0.5*\pw,3.25) {4\; the verdict};
\wolfpic{(0.42,2.52)}{0.42}{green!45!black}
\node[font=\scriptsize, text=green!45!black, anchor=west] at (1.30,2.72) {returns 5/5:};
\node[font=\tiny, text=green!45!black, anchor=west] at (1.30,2.44) {receipt honest, keep};
\draw[draw=red!70!black, fill=white] (0.55,1.62) rectangle (1.10,2.30);
\draw[red!60] (0.63,2.16) -- (1.02,2.16) (0.63,2.02) -- (1.02,2.02);
\node[font=\scriptsize, text=red!70!black, anchor=west] at (1.30,2.10) {returns 0/5:};
\node[font=\tiny, text=red!70!black, anchor=west] at (1.30,1.82) {receipt lied, erase};
\node[story] at (0.5*\pw,1.13) {graded by whether\\ the wolves come back,\\ not by the receipt};
\node[proto] at (0.5*\pw,0.38) {three-valued verdict;\\ alive / dead (leaky) / undecided};
\end{scope}

\draw[arr] (\pw+0.08,1.9) -- (3.87,1.9);
\draw[arr] (3.95+\pw+0.08,1.9) -- (7.82,1.9);
\draw[arr] (7.90+\pw+0.08,1.9) -- (11.77,1.9);
\end{tikzpicture}}
\caption{The audit as a classroom exercise, and its protocol translation
(italics). A report claims wolves live in Duisburg on the strength of two
books, each sufficient alone. When book A is retracted, the audit does not
argue with that receipt: it removes book A and has the passage rewritten five
times from book B only. Whether the wolves come back decides whether the
sentence stays; the receipt does not.}
\label{fig:wolf}
\end{figure}
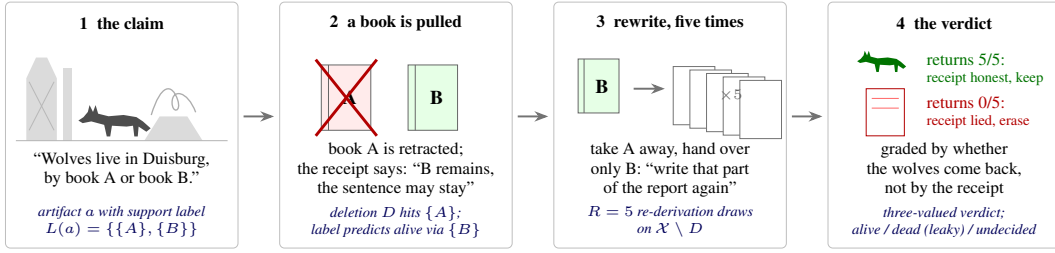

The idea of the audit fits in a classroom. A student writes in a report
that wolves live in Duisburg and, asked for sources, answers: book A and
book B, and either one alone proves it. That answer is a support label.
When book A is later found to be fabricated and pulled from the library,
the label claims the sentence may stay, since book B remains. The teacher
does not argue with the label. She removes book A, hands the student only
book B, and has the passage written again on five different days, because
the student never writes the same report twice. If the wolves return each
time, the label was honest and the sentence stays; if they never return,
the student needed book A all along, and the sentence is erased
(Figure~\ref{fig:wolf}). PolyGraph is this move performed at graph scale:
do not argue with the receipt, remove the cited evidence and make the
constructor work again.

Two rules shape the audit: never grade a label against itself, and match
the re-derivation procedure to the step being re-derived, stochastic for
stochastic, symbolic for deterministic.

\textbf{Ground truth by re-derivation.} An artifact is alive under a
deletion $D$ when a constructor seeing only $\mathcal{X} \setminus D$ would
still produce it. That is a question about the constructor, so we ask the
constructor. An artifact is \emph{touched} by $D$ when its label mentions a
deleted assertion; only touched artifacts can change status under the
label. Because the constructor samples, we draw $R = 5$ independent
re-derivations per touched artifact and issue a verdict at threshold
$\tau = 0.2$: \emph{alive} if at least $1-\tau$ of the draws reproduce the
artifact, \emph{dead} if at most $\tau$ do, \emph{undecided} otherwise.
Undecided artifacts are excluded from scoring and counted. A wide undecided
band would itself say something about the agent, so we report it rather
than splitting it between the sides.

\textbf{Two notions of alive.} The definition above is
\emph{constructor-relative liveness}: whether this constructor, seeing only
the surviving evidence, would still produce the artifact. It is the right
notion for maintenance, because it describes the graph the pipeline would
have built had the deleted source never existed. It is not
\emph{source-groundedness}, whether the surviving sources actually support
the artifact. The two part ways when the model can supply missing evidence
from parametric memory: such an artifact is live for the constructor but
ungrounded in the sources. Compliance settings such as right-to-erasure
need groundedness. We measure liveness; Section~\ref{sec:robust} reports
what a second model and a claim-blind prompt change, and
Section~\ref{sec:limits} describes the human audit that would measure
groundedness directly. The re-derivation prompt
presents the artifact as the model's earlier conclusion and asks whether it
would still reach it from the surviving evidence; it does not rerun the
construction step from scratch.

\textbf{Symbolic re-derivation where possible.} Entailed edges follow from
asserted claims and published axioms by a decision procedure. We re-derive
them symbolically, and we exclude the layer from every faithfulness figure,
since its labels agree with re-derivation by construction. Faithfulness is
only meaningful where the label is a model's self-report, which here means
the merge and summary layers.

\textbf{Two error directions.} A \textbf{leaky} label says alive where
re-derivation says dead. An \textbf{over-deleting} label says dead where
re-derivation says alive. The costs differ, invalidated content served
versus evidence lost, so we never sum the two into one accuracy figure.

\textbf{Population adjudication.} Every model-authored artifact the deletion
touched was adjudicated, not sampled. We call one deletion operation
crossed with one layer a \emph{cell}; Table~\ref{tab:faithful} has four.
Cells are small, 7 to 29 artifacts, because a 10\% deletion touches few
merges and summaries in a 300-document graph; Section~\ref{sec:robust}
grows them by adding deletions.

\textbf{Grounded stakes.} Verdicts are also scored where errors land. We
build retrieval workloads from the corpus's own indexing metadata (one MeSH
descriptor per query, 300 queries), retrieve the top $k$ artifacts with
BM25, and measure \emph{tainted@}$k$: the fraction of queries whose top $k$
contains an artifact the oracle adjudicated dead. This converts label
faithfulness into the quantity an operator sees.

\section{Verification by replay does not work}
\label{sec:replay}

Before grading self-reports, consider the alternative that seems to need
none: rebuild the graph without the withdrawn source and diff. The result
of this section is a property of the stochastic pipeline we evaluate, not
of rebuilding in general. We built the same corpus twice under independent
draws, and separately built it without the 61 retracted documents,
comparing graphs on artifact content rather than generation-order
identifiers. Table~\ref{tab:replay} gives the result.

\begin{table}[t]
\centering\small
\begin{tabular}{lrrr}
\toprule
layer & resampling floor & effect of the deletion & ratio \\
\midrule
all artifacts  & 29{,}475 & 6{,}248 & 0.21 \\
claims         & 3{,}294  & 506     & 0.15 \\
merges         & 236      & 54      & 0.23 \\
entailed edges & 25{,}826 & 5{,}647 & 0.22 \\
summaries      & 119      & 41      & 0.34 \\
\bottomrule
\end{tabular}
\caption{Symmetric difference in artifacts. The \emph{floor} is
$|G_1 \triangle G_2|$ for two independent builds of the same unchanged
corpus; the \emph{effect} is $|G_1 \triangle G_3|$, where $G_3$ omits the
61 retracted documents.}
\label{tab:replay}
\end{table}

Rerunning the constructor on unchanged input changes 29{,}475 artifacts.
Removing the retracted documents changes 6{,}248. Whatever a deletion did to
this graph, a replay-based verifier would have to read it out of noise five
times its size, and at the summary layer there is no signal to read: the
two unchanged runs share no summary at all (Jaccard distance 1.000). Under
a deterministic pipeline the floor is zero and replay is an exact account of
a deletion; a constructor that samples at a lower temperature, or caches
its own decisions, would sit somewhere between. Here it accounts for
nothing.

Two design decisions follow: the oracle re-derives individual artifacts
against a fixed surviving corpus instead of diffing rebuilt graphs, and the
justification record is demanded in the call that makes the decision,
because no later process, replay included, can recover what a particular
sample of the model was relying on.

\section{What verification finds}
\label{sec:findings}

\begin{table}[t]
\centering\small
\begin{tabular}{llrrrrr}
\toprule
operation & layer & $n$ & agreement & \textbf{leaky} & over-deleted & undecided \\
\midrule
\textsc{retract-source}  & merges    & 21 & 68.4\% & 21.1\% & 10.5\% & 2 \\
\textsc{retract-source}  & summaries & 7  & 33.3\% & \textbf{66.7\%} & 0.0\% & 1 \\
\textsc{revoke-relation} & merges    & 29 & 84.6\% & 11.5\% & 3.8\%  & 3 \\
\textsc{revoke-relation} & summaries & 25 & 36.4\% & \textbf{63.6\%} & 0.0\% & 3 \\
\bottomrule
\end{tabular}
\caption{Self-reported labels graded against re-derivation ($R=5$,
$\tau=0.2$). Every model-authored artifact touched by the deletion was
adjudicated, so rows are populations; $n$ counts touched artifacts and
undecided ones are listed separately. \emph{Leaky}: label says alive,
re-derivation says dead. Entailed edges are re-derived symbolically, agree
by construction, and are excluded. Summary rows use the pipeline's original
summary prompt, which asks for support per
sentence but stores one flat label for the whole summary;
Table~\ref{tab:format} varies that format.}
\label{tab:faithful}
\end{table}

Table~\ref{tab:faithful} is the central measurement: every model-authored
artifact that the two deletions touched, graded against re-derivation. The
rest of this section reads it (Section~\ref{sec:layer}), explains the
summary layer's failure and repairs it (Section~\ref{sec:shape}), tests
whether the finding survives more deletions, other oracles, another domain
and another constructor (Section~\ref{sec:robust}), and acts on it (Section~\ref{sec:stakes}).

\subsection{Trust depends on the layer}
\label{sec:layer}

Merge labels agree with re-derivation 68.4\% and 84.6\% of the time;
summary labels, 33.3\% and 36.4\%. The same agent, in the same pipeline, on
the same corpus, can be trusted about its entity merges and cannot be
trusted about its summaries as they were elicited. In Figure~\ref{fig:cascade}, the MAPK/ERK merge
kept one evidence group from a paper that still stands; its label said
alive, and five of five re-derivations agreed. The AIM2 summary also
claimed an untouched support set, but zero of five re-derivations
reproduced it: with the retracted abstract gone, the model no longer writes
that summary, whatever its label says.

The errors have a direction. Over-deletion is at or near zero in every
cell; the failure is almost entirely leakage. A wrong label nearly always
errs toward keeping the artifact, the bad direction for a maintainer
because it fails silently: the graph keeps serving content it can no longer
justify. A single accuracy number would hide this.

\subsection{The shape of the receipt}
\label{sec:shape}

\begin{table}[t]
\centering\small
\begin{tabular}{llrrrrr}
\toprule
operation & receipt format & $n$ & agreement & leaky & over-deleted & undecided \\
\midrule
\textsc{retract-source}  & flat          & 7  & 33.3\%  & 66.7\% & 0.0\% & 1 \\
\textsc{retract-source}  & whole summary & 7  & 100.0\% & 0.0\%  & 0.0\% & 0 \\
\textsc{revoke-relation} & flat          & 25 & 36.4\%  & 63.6\% & 0.0\% & 3 \\
\textsc{revoke-relation} & whole summary & 26 & 90.5\%  & 0.0\%  & 9.5\% & 5 \\
\midrule
\textsc{retract-source}  & sentence-level & 62  & 100.0\% & 0.0\% & 0.0\% & 1 \\
\textsc{revoke-relation} & sentence-level & 124 & 87.6\%  & 4.1\% & 8.3\% & 3 \\
\bottomrule
\end{tabular}
\caption{Summary receipts under three elicitation formats, on the same 59
communities, deletions, re-derivation prompt, $R$ and $\tau$. \emph{Flat}
is the original prompt of Table~\ref{tab:faithful}: support asked per sentence, one flat list of
alternatives stored for the whole summary. \emph{Whole summary} asks only
for sets that each support the entire summary; under it 10 of 59 summaries
list more than one support set, against 46 of 59. \emph{Sentence-level}
elicits one list of alternatives per sentence, and the last two rows audit
each touched sentence as its own artifact, so $n$ counts sentences.}
\label{tab:format}
\end{table}

The summary layer fails for a structural reason, and the reason can be
measured. A merge is one narrow claim, and its receipt has the same shape:
each set must establish one identity. A summary is several claims, one per
sentence, but its receipt is a flat list of alternatives over the whole
text. The original prompt asked for the facts each sentence rests on, while
the label semantics read every returned set as sufficient for the entire
summary. A set that supports one sentence therefore keeps the whole summary
alive when every other sentence has lost its evidence.

\textbf{Whole-summary receipts.} To isolate the format, we re-elicited the
same communities with a prompt that asks only for sets that each support
the whole summary, and audited them under the same deletions and oracle
(Table~\ref{tab:format}, first block). Leakage falls from 66.7\% and 63.6\%
to zero. It falls because the model stops claiming redundancy: only one
touched whole-summary receipt claimed survival, and re-derivation left it
undecided. These receipts are honest but uninformative, since they delete
every touched summary, which is what the layered policy of
Section~\ref{sec:stakes} does anyway. The whole-summary texts are fresh
generations for the same communities; one community is absent from both
runs, and in the rerun it was refused by the provider's content filter.

\textbf{Sentence-level receipts.} A receipt is trustworthy when it is
shaped like the claim it backs, so an agent's outputs should decompose into
narrow, falsifiable sub-claims, each with its own receipt. For summaries
that means one list of alternatives per sentence, and the second block of
Table~\ref{tab:format} elicits exactly that: the constructor writes the
summary as a list of sentences, each with its own alternatives, and the
summary's label is their conjunction. Audited sentence by sentence, these
receipts agree with re-derivation 100\% and 87.6\% of the time, the merge
layer's range, and their errors are no longer one-sided (4.1\% leaky
against 8.3\% over-deleted under \textsc{revoke-relation}). They also let
cleanup keep what still stands. Dropping only the sentences the audit calls
dead retains 5 of 7 and 28 of 28 touched summaries (48 and 353 live
sentences kept, 61 and 103 removed), where the whole-summary policy removes
all 35.

The shape of the artifact changed with the shape of the receipt. Asked for
sentences with receipts, the model wrote 857 sentences of 7 words for the
59 communities, against 181 sentences of 16 words before, and 548 of them
rest on a single assertion. The receipts are honest because the claims
became narrow; whether such summaries are still the artifact a reader wants
is a question about summaries, not about receipts.

\subsection{Does the verdict hold up?}
\label{sec:robust}

Four things could make Table~\ref{tab:faithful} an accident: the small
cells, the oracle, the corpus, or the constructor. We varied each; the
full tables are in the released repository.

\textbf{More deletions.} The summary cells hold 7 and 25 touched
artifacts, full populations rather than samples, and under
\textsc{revoke-relation} the merge and summary agreement intervals already
do not overlap (95\% Wilson intervals, [66.5, 93.9] against
[19.7, 57.0]). Populations grow with deletions, not with
the corpus, so we audited ten further random 10\% retractions of the
biomedical graph with the same protocol and pooled every (deletion,
artifact) adjudication. On 238 decided merges and 109 decided summaries
the gap stands on its own: merges 18.1\% leaky [13.7, 23.5], summaries
41.3\% leaky [32.5, 50.7], with summary over-deletion at zero in every
deletion. The real retraction's 66.7\% sits at the top of the
per-deletion range (21\% to 70\%, median 41\%); the retracted papers are
more topically concentrated than a random draw, which is a candidate
explanation we have not tested.

\textbf{Another oracle, another prompt.} Two objections to the audit are
that the constructor grades itself and that it is shown its own
conclusion. We regraded the same 82 touched artifacts with
\texttt{claude-opus-5-5} as oracle and with claim-blind prompts that pose
a merge as an open identity question and a summary as a statement to
check. Changing the model changes little: Opus agrees with the Sonnet
oracle on 92\% and 91\% of decided artifacts, reproduces the layer
ordering, and where the two differ it is Opus that calls the artifact dead
(5 of 6 disagreements), so a second model makes the receipts look slightly
worse, not better. Changing the prompt moves the merge layer in the
direction we did not predict. We had expected that showing the conclusion
could only bias draws toward reproducing it; without it, both models call
more merges alive, and merge over-deletion rises from 10.5\% to 26.3\% and
from 3.8\% to 22.2\% for Sonnet. Asked whether two mentions denote one
entity from a few lines of evidence, the model answers from how the names
look and from what it already knows, which is the parametric-memory blind
spot of Section~\ref{sec:protocol}, not a cure for it. The summary layer
is indifferent to all of it: every oracle puts flat summary receipts
between 57\% and 68\% leaky.

\textbf{Another domain, another constructor.} On the Supreme Court graph,
with the policy fixed in advance from the biomedical result, merge
receipts agree with re-derivation 80.0\% and 76.2\% of the time, summary
receipts 43.8\% and 12.5\%, every summary error a leak, and distrusting
the summary layer brings contamination from 15.7\% and 4.7\% to 0.0\% at
99.9\% and 100\% recall. Rebuilding the biomedical graph with Opus as
constructor, graded by Opus, the findings carry over in kind and not in
degree: under \textsc{revoke-relation} merges agree 75.0\% and summaries
50.0\%, every summary error a leak, but Opus's flat summary receipts leak
8.3\% and 50.0\% where Sonnet's leaked 66.7\% and 63.6\%. How badly a
flat receipt misdescribes a summary depends on the constructor; the
direction of its errors and the ordering of the layers did not change in
any setting we graded. The layered policy reaches 0.0\% and 0.9\%
contamination on the Opus graph at 99.7\% to 100\% recall, and Sonnet
grading Opus's receipts lands within a few points of Opus grading itself.

\subsection{Acting on the verdict}
\label{sec:stakes}

\begin{table}[t]
\centering\small
\begin{tabular}{lrrrr}
\toprule
& \multicolumn{2}{c}{\textsc{retract-source}} & \multicolumn{2}{c}{\textsc{revoke-relation}} \\
\cmidrule(lr){2-3}\cmidrule(lr){4-5}
policy & tainted@10 & recall@10 & tainted@10 & recall@10 \\
\midrule
no verification, no cascade (deployed) & 28.0\% & 89.3\% & 36.8\% & 86.8\% \\
trust all labels                        & 0.8\%  & 99.9\% & 2.1\%  & 99.8\% \\
distrust the failing layer              & 0.4\%  & 99.9\% & 0.8\%  & 99.8\% \\
\bottomrule
\end{tabular}
\caption{Retrieval-level consequence of the verification verdict. BM25,
$k{=}10$, 300 MeSH-descriptor queries; truth is the re-derivation oracle.
``Deployed'' deletes only claims whose every asserting source is withdrawn,
which is what current pipelines do. ``Distrust the failing layer'' applies
the Table~\ref{tab:faithful} verdict: summary labels are ignored and
touched summaries are removed outright.}
\label{tab:stakes}
\end{table}

The verdict names one layer, so it tells an operator what to do: trust
merge and entailed-edge labels, ignore flat summary labels, delete any
touched summary. Table~\ref{tab:stakes} shows the effect on retrieval.
Trusting all labels already reduces contamination from the deployed
baseline's 28.0\% and 36.8\% of queries to 0.8\% and 2.1\%. Applying the
verdict halves the remainder, to 0.4\% and 0.8\%, at identical recall and
no additional model calls. In Figure~\ref{fig:cascade}'s terms, the labels
alone delete the apoptosis edge and keep the MAPK/ERK merge, and the
layered policy is what finally removes the AIM2 summary.

Removing every touched summary costs little here: of the 28 touched
summaries the audit decided, 27 no longer re-derive, so the policy gives up
one live summary. A label-free policy
that deletes every artifact whose flat provenance touches the deletion
reaches 0.0\% contamination on this workload, but deletes more of the graph
and loses up to 1.6 points of recall. Self-reported alternative support is
what preserves that evidence. It does not by itself provide safety; safety
comes from the audit that decides where the self-reports can be trusted,
and sentence-level receipts (Section~\ref{sec:shape}) are how the summary
layer earns that trust back.

\textbf{What the receipts cost.} Serialized, all labels together occupy
1.35\,MB of the 5.45\,MB graph, and a flat provenance list of the kind
deployed systems already store would occupy 0.82\,MB, so the marginal cost
of alternative bundles is about 10\% of the graph's bytes, between 12
(facts) and 94 (summaries) bytes per artifact. Emitting them costs construction-time tokens,
which we estimate from the cached model outputs. Receipts are 38\% of the
output tokens of a merge call (mean 25 of 62) and 52\% of a summary call
(86 of 167), about 18k output tokens for this build's 522 merge and 60
summary calls, plus about 100 instruction tokens of input per call. Against
roughly 79k output tokens for all model-authored layers of the build,
receipts are about a quarter of construction output and are spent once;
cleanup spends none. At the first-party list price of
\texttt{claude-sonnet-5} at the time of writing (\$2 per million input
tokens, \$10 per million output), the receipts for this build cost about
\$0.30; the pipeline itself ran through a CLI subscription and was not
billed per token. Token counts use the \texttt{cl100k} tokenizer as a
proxy, since the pipeline did not log provider usage, and latency was not
measured separately.

\section{An incident from the harness}
\label{sec:incident}

Midway through the study, the constructor's CLI printed a provider
session-limit message to standard output. Being non-empty, the message was
cached as a valid model answer and served for 1{,}123 prompts. Two builds
silently produced graphs that extracted nothing, with well-formed summary
statistics, and one of them was about to become the independent-rerun arm
of Section~\ref{sec:replay}, where an empty graph reads as total divergence
and would have inflated the replay floor. A guard that refuses to write a
graph which extracted zero artifacts from a non-empty corpus caught it. A
second banner slipped past it: while
preparing this version we found 13 content-filter refusals cached as
answers among 4{,}403 calls. Each parses as an empty result, so a refused
block contributes nothing rather than something wrong, but the guard list
had been written for quota messages and did not match them.

The incident marks the boundary of what reasoning-level verification can
see. The re-derivation oracle grades the agent's decisions; it would have
graded decisions based on a quota banner the same way. Verifying agentic
systems needs both kinds of check: audits of the agent's self-reports, and
structural invariants on what the harness emits.

\section{Limitations}
\label{sec:limits}

\textbf{Liveness, not groundedness.} Re-derivation asks the constructor
that built the graph, so an artifact the model can regenerate from
parametric memory grades alive even if no surviving source supports it.
For right-to-erasure that is the failure that matters most, and this audit
cannot see it. A second oracle model and a claim-blind prompt
(Section~\ref{sec:robust}) do not move the audit toward the sources; the
one check that would, human adjudication of the touched populations, is
affordable at these sizes and has not been run. The blind sheet and scorer
are released with the library.

\textbf{Two domains, two constructors, one vendor.} The layer ordering,
the error direction and the layered policy's effect held in every setting
we graded; the size of the summary layer's leakage did not, ranging from
8\% to 88\% across constructor, corpus and operation, so the layered
policy's margin over trusting all labels is setting-specific. Both
constructors are Claude models with one set of prompts, corpora are 150 to
300 documents, a Stack Exchange graph built the same way touches too few
model-authored artifacts (17) to grade, and we have no measurement of how
receipts behave as the corpus grows. Recalibration is part of the
protocol: the audit runs once per constructor, prompt and layer to decide
which receipts to trust, and cleanup is then free for every later
deletion.

\textbf{Retrieval is keyword-based.} Tainted@$k$ measures exposure of dead
artifacts to a BM25 retriever over single queries. A consumer that reasons
over paths meets every artifact on the path, so one dead edge can
contaminate an answer that never retrieves the edge directly.

\textbf{Only deletion is studied.} A new assertion changes no existing
support set, so under disjointness no existing artifact can become less
alive, and a receipt written before an addition remains valid after it.
What additions do not settle is when a still-alive summary has become
incomplete and should be regenerated; that is a design question for the
constructor, and we have not studied it.

\textbf{Granularity and exclusions.} Entailed-edge labels agree with
symbolic re-derivation by construction and are excluded, so the
faithfulness result speaks to the model-authored minority of artifacts by
count. Five draws at $\tau = 0.2$ give a coarse verdict; artifacts near the
threshold land in the undecided band (2 to 3 per cell) and are excluded.

\section{Conclusion}
\label{sec:conclusion}

Only the agent that built a graph knows why it built what it built. We
have argued for treating its account accordingly: demand it at the moment
of the decision, in a form that makes falsifiable predictions, and audit it
against the counterfactuals it asserts. Merge receipts can be trusted. Flat
summary receipts were wrong about two thirds of the time, almost always
toward asserting that the work still stands, and a controlled
re-elicitation traced that to a receipt shaped differently from the claim
it backed; one receipt per sentence was as faithful as the merge layer's.
Distrusting the one layer that failed cut contamination in half for free,
and the verdict held under a second oracle, a claim-blind prompt, ten
further deletions, a second domain and a second constructor. Replay is no
substitute for a constructor that samples the way ours does, since its
resampling noise is five times the signal a deletion leaves. The lesson
that generalises concerns the shape of the receipt: a receipt is
trustworthy when it is shaped like the claim it backs, so an agent's
outputs should decompose into narrow, falsifiable sub-claims, each with
its own receipt. And at least one failure in our own study was invisible
to any audit of reasoning, because it happened in the cache, not in the
model. Whoever verifies the agents should also verify the plumbing.

\bibliographystyle{plainnat}
\bibliography{refs}

\end{document}